\documentclass[letterpaper, 10 pt, conference]{ieeeconf}  
\IEEEoverridecommandlockouts
\usepackage{graphicx}
\usepackage{epstopdf}
\usepackage{amsmath}
\usepackage{amssymb}
\usepackage{subfigure}

\usepackage{multirow}
\usepackage{pbox}
\usepackage{cite}
\usepackage{algorithm}
\usepackage[noend]{algpseudocode}
\usepackage{booktabs}
\usepackage{wrapfig}
\usepackage{lscape}
\usepackage{bm}
\usepackage{url}
\usepackage{esvect}
\usepackage{xcolor,soul}
\usepackage{txfonts}
\usepackage[normalem]{ulem}
\usepackage{tikz}
\usepackage{float}
\usepackage{pifont}
\usetikzlibrary{matrix,calc}

\usepackage{cuted}   
\usepackage{capt-of}

\DeclareSymbolFont{extraup}{U}{zavm}{m}{n}
\DeclareMathSymbol{\varheart}{\mathalpha}{extraup}{86}
\DeclareMathSymbol{\vardiamond}{\mathalpha}{extraup}{87}

\DeclareMathOperator*{\argmax}{arg\,max}
\DeclareMathOperator*{\argmin}{arg\,min}

\algnewcommand\algorithmicforeach{\textbf{for each}}
\algdef{S}[FOR]{ForEach}[1]{\algorithmicforeach\ #1\ \algorithmicdo}

\renewcommand{\baselinestretch}{1.0}

\begin{document}

\author{Jonathan Reasoner and Nicola Bezzo
\thanks{Jonathan Reasoner and Nicola Bezzo are with the Department of Electrical and Computer Engineering, University of Virginia, Charlottesville, VA 22904, USA. 
Email: {\tt \{vqh7rx, nb6be\}@virginia.edu}}\\\vspace*{-3cm}}

\title{\LARGE \bf Higher Order Reasoning for Collaborative Communicationless \\Mobile Robot Operations}
\maketitle

\renewcommand{\baselinestretch}{0.925}

\maketitle

\vspace{-1cm}
\begin{strip} 
\begin{center}
\vspace{-1.95cm} 
\includegraphics[width=0.9\linewidth]{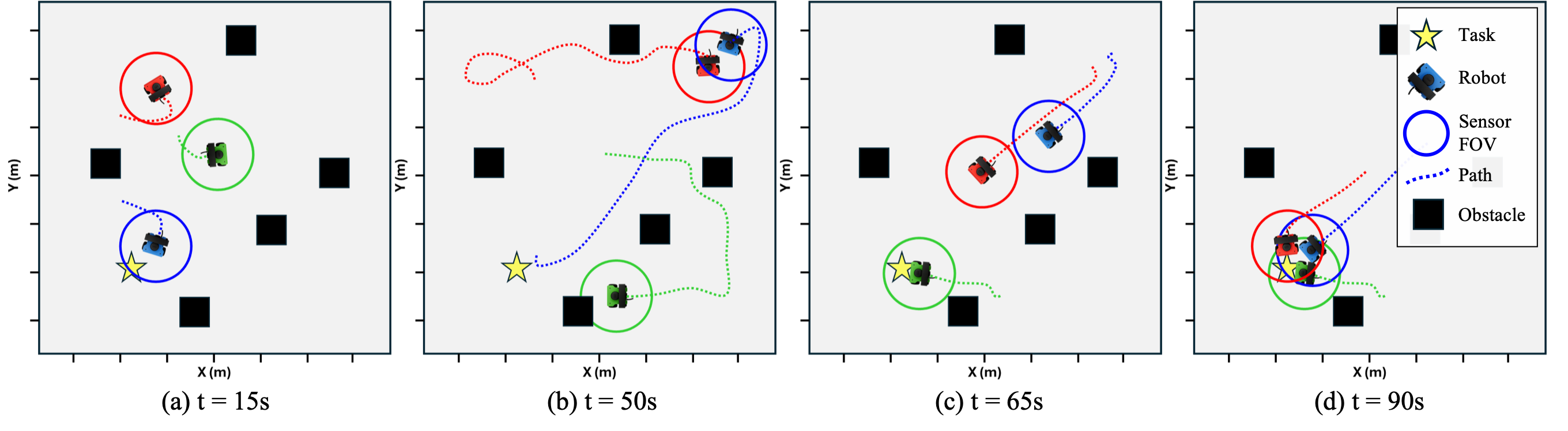}
\vspace{-12pt}
\captionof{figure}{ Illustration of the proposed higher order reasoning for a 3 robot operation in a communicationless environment. {\em Blue} finds a task (a) and decides to go fetch {\em red} (b) because it reasons that {\em red} will take longer than {\em green} to find the task with its current plan. Meanwhile, {\em green} finds the task (c) and reasons that {\em blue} likely found it earlier and went to fetch {\em red}, thus triggering it to remain at the task to wait for the rest of the team (d).}
\label{fig:motivation}
\vspace{-18pt}
\end{center}
\end{strip}
    








\begin{abstract}

In communicationless environments, multi-robot systems must operate without the constant information exchange that many coordination strategies typically assume. This paper presents a novel dynamic epistemic planning framework that enables implicit coordination and long horizon planning through higher-order reasoning among robots. With our approach, robots form and propagate higher-order belief particles, update world beliefs using Bayesian inference, and select actions via a behavior tree that anticipates teammates’ likely decisions. A temporally aware  Model Predictive Path Integral (MPPI) controller integrates this reasoning into low-level execution, allowing robots to plan intercepts and adapt trajectories under partial observability. The proposed framework is evaluated in both simulations and physical experiments, where it consistently reduces task completion time compared to a first-order baseline, demonstrating that epistemic logic can serve as a robust foundation for resilient coordination in communication-restricted domains.
\end{abstract}



\section{Introduction}\label{sec:intro}

Multi-robot systems (MRS) provide advantages over single-robot systems for many task-oriented applications such as search and rescue, disaster relief, surveillance, and reconnaissance \cite{uses_cite}. In such operations, task objectives are often unknown \textit{a priori} and may require multiple robots to complete; therefore, deploying an MRS to localize these tasks is beneficial.
However, MRS effectiveness often relies on unrealistic assumptions of constant communication \cite{jamming} or carefully planned rendezvous strategies \cite{bramblett_limited_connectivity}. With these considerations in mind, this paper focuses on answering the following question: {\em How can we effectively coordinate an MRS to perform collaborative operations without explicit communication?} While this is a very strict constraint, we note that communication is often the bottleneck in MRS coordination, presenting delays and data loss. Additionally, communication may be not possible due to range constraints, undesired as in military scenarios where detection is unwanted, or denied due to jamming or bandwidth congestion.

To this end, we propose a Theory of Mind (ToM)-based epistemic planning method that empowers each robot in the MRS with the ability to perform higher-order reasoning and predict and plan actions using the ``I know that you know that I know" empathetic paradigm found in social/cognitive science. This approach is inspired by how humans form beliefs, reason about higher-order knowledge, and empathize with others’ perspectives to reach a consensus, enabling collaboration. 
ToM hypothesizes that through the combination of such belief and empathy modeling, humans can predict the behavior of others, adapt their own plans accordingly, and coordinate effectively, even in the absence of communication.

In this work, we extend the concept of ToM to autonomous systems, enabling robots to reason over both what they know and what they believe other robots may know. We will show that coordination strategies based on higher-order reasoning provide a robust basis for persistent MRS operations.
Fig. \ref{fig:motivation} illustrates the problem and outcome from our proposed ToM-based epistemic planning method. In this example, taken from the simulation, three robots are deployed to find and perform a task that requires the entire team to complete. The task location is unknown a priori. Using the proposed approach, the three robots begin by exploring the environment without communication using a frontier method. The {\em blue} robot finds the task, observes that it requires the rest of the team and evaluates what behavior to perform: 1) waiting to perform the task, 2) continuing exploring, or 3) seeking help. By predicting the behavior and future states of the other teammates, {\em blue} decides to proceed with 3) and fetch {\em red} who will take longer than {\em green} to find the task. {\em Blue} reasons that it can shorten the time for {\em red} to reach the task and complete the entire mission by selecting this behavior. While {\em blue} is fetching {\em red}, {\em green} discovers the task and reasons about what may have happened in the past and what the likely behaviors of the other robots are. {\em Green} ``rewinds'' time and predicts that {\em blue} may have found the task and it may have gone to fetch {\em red}, thus it chooses to wait at the task where it expects to see {\em blue} and {\em red} approaching in a near time horizon. {\em Blue} can similarly predict the expected time that it takes for {\em green} to reach the task and {\em green's} behavior after task discovery. The same applies to {\em red} when it reaches the task.
The example in Fig.~\ref{fig:motivation} highlights the core higher-order reasoning principles behind our approach that allow robots to: `think' about what has happened and what will happen over a horizon, evaluate different actions and the possible outcomes, and select the best behavior while considering the entire team's capabilities.

We achieve this implicit higher-order reasoning and coordination through the use of: 1) Dynamic Epistemic Logic (DEL) which creates and propagates belief and empathy particles, allowing a robot to model the evolution of states and behaviors over a long horizon, 2) Bayesian belief updates, which allows the fusing of local observations with previously held epistemic state beliefs, and 3) a behavior tree search process which uses previously formed belief and empathy particles to determine the optimal behavior from a set of possible behavior primitives. Additionally, we propose: 4) a modified MPPI controller that produces stochastically optimal control inputs under a non-linear, time-varying objective function to fetch other robots, and 5) a belief-informed frontier exploration process which is used for efficient exploration under communication constraints.

Our contribution is two-fold: 1) we introduce a long-horizon estimation–planning–coordination framework for non-communicative multi-robot teams that couples DEL-based higher-order reasoning with belief-informed frontier exploration and an MPPI to achieve path planning in dynamic environments with non-linear cost functions; 2) we show that leveraging higher-order reasoning enables behavior-tree–driven long-horizon plans that adapt to unexpected situations and accelerate task completion across both simulations and experiments.

The rest of the paper is organized as follows. Sec.~\ref{sec:related_work} reviews the state of the art in MRS coordination; Sec.~\ref{sec:problem} formally defines the problem; Sec.~\ref{sec:approach} presents the MRS estimation–planning–coordination framework for communicationless systems; Secs.~\ref{sec:simulations} and \ref{sec:experiments} present simulation and experimental results; and Sec.~\ref{sec:conclusion} provides conclusions and outlines future work.
\vspace{-2pt}



\section{Related Work}\label{sec:related_work}
Interest in multi-robot exploration and task allocation is rising, driven by affordable platforms and goals of reducing human involvement in hazardous or menial tasks \cite{mrs_survey}. Previous efforts have attempted to solve the coordination challenge associated with MRS through market-based approaches that leverage an auctioneer to determine the optimum agent to explore an area\cite{centralized_market_based} or through decision-theoretic approaches \cite{coordinated_multi_robot_exploration}, but these methods require assured communication between robots and the central controller. To overcome the shortcomings of centralized systems, decentralized approaches have been increasingly proposed due to their flexibility and robustness. In \cite{bramblett_limited_connectivity, coordinated_multi_robot_exploration}, global communication assumptions are relaxed and a decentralized approach is implemented that accounts for a limited communication range. While these approaches allow for more realistic environments, systems can still fail due to malfunction or intentional interference such as jamming or signal degradation \cite{jamming}. 
Prior non-communicative work have leveraged biomimicry for coordination \cite{bees}, using visual cues, predefined sensorimotor behaviors, and a priori search patterns, with simple evaluations in obstacle-free environments. 

In order to address the challenges of coordinating communicationaless MRS, we implement a ToM inspired approach. ToM focuses on how humans infer the beliefs, intentions, and desires of other humans and use that knowledge to improve our understanding of the world around us \cite{theoryofmind, ToM_kids}. To integrate Theory of Mind (ToM) into autonomous systems, epistemic planning has been proposed as a way to generate plans that consider what a robot knows and believes, as well as what it does not know \cite{gentle_intro,bolander_multi_agent}. Dynamic epistemic logic (DEL) extends epistemic logic with a framework for information change, specifying how observations transform an agent’s belief state \cite{bolander_DEL_epistemic_planning}. This approach enables higher-order reasoning: robots form beliefs about other robots, about what other robots believe about them, and about what those robots believe about third robots. Following \cite{ToM_kids}, we refer to these as first-, second-, and third-order reasoning.

Through the use of DEL-based higher-order reasoning, this work advances the state of the art in decentralized MRS coordination for unknown, partially observable environments where explicit communication is not possible.



\section{Problem Formulation}
\label{sec:problem}

    Let us consider an MRS team, $\mathcal{R}$, of $n_r$ robots operating in a partially known, communicationless environment $\mathcal{W}\subseteq\mathbb{R}^2$ in which the goal is to discover and complete a task $g$ in the shortest amount of time.

Let $\mathcal{R}_g\subseteq\mathcal{R}$ denote the subset of robots required to complete $g$. For any robot $i\in \mathcal R_g$ we define $\tilde{T}_i$ as the estimated task completion time, measured as the time lapsed from the start of the mission to the completion of $g$. With this, the total mission time $\mathcal T$ can be written as:
\begin{equation}
\label{eq:missionT}
\mathcal{T}=\mathbb{E}\left[\max_{i\in \mathcal{R}_g}\tilde{T}_i,\right].
\end{equation}
We can then formally define our problem as:

\textbf{Problem 1: {\em Communicationless Collaborative Task Discovery and Completion}.}
Design a strategy to dynamically select the best motion planning policy $\pi_i\in\Pi$ for each robot $i \in \mathcal{R}$ such that $g$ is discovered and completed in the shortest amount of time possible:
\begin{equation}
\label{eq:policy_selection}
\pi_i = \argmin_{\pi \in \Pi}\mathcal{T} \quad \forall i \in \mathcal{R}.
\end{equation}

As we will show next, by leveraging higher-order reasoning and planning - which for brevity we call epistemic planning - paired with future and backward horizon predictions, each robot is able to minimize this global cost by selecting behaviors that others can predict in the absence of explicit direct communication. 
\vspace{-4pt}

\section{Approach}\label{sec:approach}
\begin{figure}[t]
\centering
\includegraphics[width=0.48\textwidth]{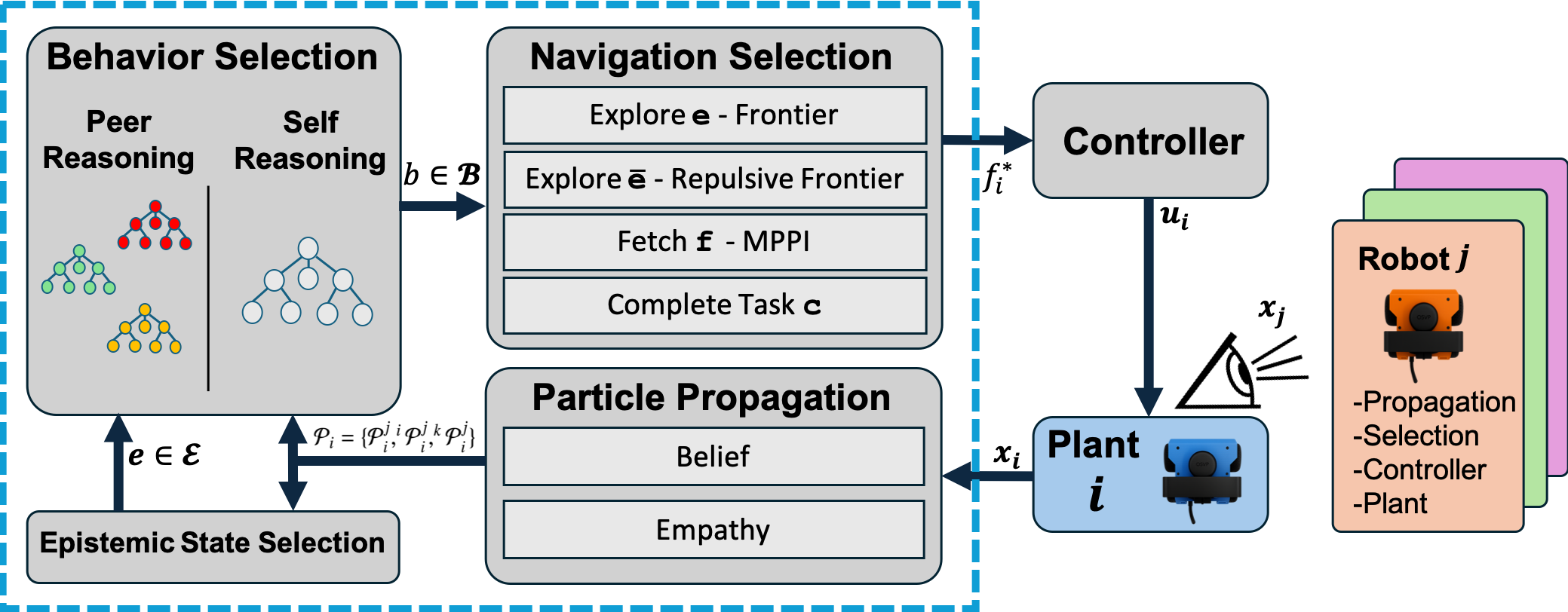}
\vspace{-15pt}
\caption{Diagram of our proposed approach. Our contribution is shown in the blue dashed box.}
\label{fig:system_overview}
 \vspace{-15pt}
\end{figure}

Our approach is summarized in Fig. \ref{fig:system_overview}. Each robot begins by exploring the environment in search of tasks using a Frontier method \cite{frontiers}. Once a robot discovers a task, it performs an epistemic belief assessment, predicting the likely behaviors of other teammates by propagating forward and backward belief particles. These particles inform the behavior tree, which applies high-level reasoning to evaluate all actions and predict associated outcomes to best accomplish the task. The optimal behavior is then selected, and the process continues until all tasks are completed or the environment has been fully explored. We now present how each component of the approach is implemented, starting with a preliminaries section. We then discuss how belief and empathy particles are created and propagated to predict teammates' behaviors and select the optimal behavior to execute.


\subsection{Preliminaries}
\label{sec:preliminaries}
\subsubsection{Notation}
Throughout the paper, predicted quantities are denoted by a tilde ($\tilde{x}$), unit vectors by a hat ($\hat{\bm{v}}$), and vectors are boldface ($\bm{x}$).
The state vector, $\bm{x}_i(t) = \big[x_i(t),y_i(t),\theta_i(t)\big]^\top\in \mathbb{R}^2 \times \mathbb{S}$, consists of the $(x,y)$ position and orientation, $\theta$ for robot $i$.
Lastly, the unit bearing between two vectors is denoted as: 
\begin{equation}
\label{eq:unit_vector}
\hat{\bm{d}}(\bm{x}_a, \bm{x}_b) \;=\; \frac{\bm{x}_b-\bm{x}_a}{\lVert \bm{x}_b-\bm{x}_a\rVert}.
\end{equation}
\subsubsection{Dynamic Epistemic Logic (DEL)}
DEL is a reasoning framework that allows robots to translate observations into beliefs, considering uncertainty and incompleteness of information, and reason over those beliefs for planning in systems with distributed knowledge and capabilities \cite{gentle_intro}.

In order to use DEL in our approach, we construct 3 vectors which hold information that robot $i$ has attained through observations and reasoning, specifically: the robot's pose $\bm{x}_i$, the epistemic state $\bm{e}_i$, and the behavior state $\bm{b}_i$. 

The epistemic state, $\bm{e}_i(t) \in \mathbb{R}^{n_r}$ encodes team knowledge with entries $e_i^{j}(t)\in\{0,1\}$, where $e_i^{j}(t)=1$ if robot $i$ believes $j$ has discovered the task. For our work, we only consider environments with a single task, but our approach can be expanded to $n_g$ tasks by expanding the epistemic state to $\bm{e}_i(t) \in \mathbb{R}^{n_r \times n_g}$.
The behavior state, $ \bm{b}_i(t)=\bigl[ \tilde{b}_i^0(t), \dots, \tilde{b}_i^{n_r}(t)\bigr] $, describes the predicted behavior of all robots within $\mathcal{R}$.  



For collaborative multi-robot operations, most required behaviors can be decomposed into a set of behavior primitives. For this work, we define these primitives as i) {\em explore} ($\texttt{e}$): exploring an area in search of the task, ii) {\em complete task} ($\texttt{c}$): going to an identified task or waiting for other robots to arrive at the task, iii) {\em fetch} ($\texttt{f}_\texttt{j}$): deliberately creating an interaction with another robot, $j$ with the intention of driving it towards the task, or iv) {\em modified explore} ($\bar{\texttt{e}}$): a temporarily modified search pattern resulting from an unexpected interaction with another robot. This modified search pattern drives a robot away from a specific robot before resuming normal frontier exploration, $\texttt{e}$, behavior. With this minimal set of behavior primitives, each robot can select a policy according to \eqref{eq:policy_selection} for the class of collaborative tasks addressed herein.

\subsection{Belief and Empathy Formulation and Propagation} 
\label{sec:epistemic_belief}

Each robot $i$ starts the operation in exploration behavior $b_i=\texttt{e}$ and with an epistemic state $e_i=0$ until a task is discovered in which case $e_i=1$. Within our epistemic framework, each robot propagates belief states $\bm x$ for all robots over a predictive time horizon  $H$. In this way, each robot $i$ can plan according to its belief about other robots and empathize with where other robots expect $i$ to be located and what behavior they expect $i$ to be doing, even without communicating. We define a set of particles $\mathcal P_i=\{ \mathcal{P}_i^{j}, ^i{\mathcal{P}}_i^{j}, ^k{\mathcal{P}}_i^{j}\}$ to represent these belief and empathy states from the perspective of $i$, where: 
\begin{itemize}
\item $\mathcal{P}_i^{j}(t) = \Bigl\{ \tilde{\bm{x}}^j_i(t) | \tilde{b}_i^j\Bigr\}_t^{H}$ is the first order belief indicating $i$'s estimated state of $j$ given predicted behavior of $j$ over a time horizon $H$;
\item $^i{\mathcal{P}}_i^{j}(t) =  \Bigl\{\,^i\tilde{\bm{x}}^{j}_i(t) | ^i\tilde{b}_i^{j}\Bigr\}_t^{H}$ is the second-order empathy that indicates what $i$ believes that $j$ believes about $i$; and 
\item $^k{{\mathcal{P}}}_i^{j}(t) = \ \Bigl\{\,^k\tilde{\bm{x}}^{j}_i(t) | ^k\tilde{b}_i^{j}\Bigr\}_t^{H}$ represents what $i$ believes that $j$ believes about $k$.
\end{itemize}
These particles are used to inform robot $i$'s frontier exploration strategy and higher-order reasoning.

\begin{figure}[!t]
    \centering
    \subfigure[Frontier selection process at $t$]{\includegraphics[width=0.47\linewidth]{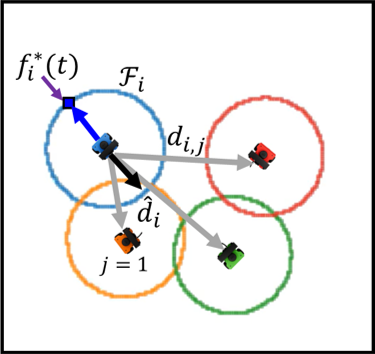}\label{fig:frontier_1}}%
    \vspace{-3pt}
    \hfill
    \subfigure[Frontiers at $t+10$]{\includegraphics[width=0.47\linewidth]{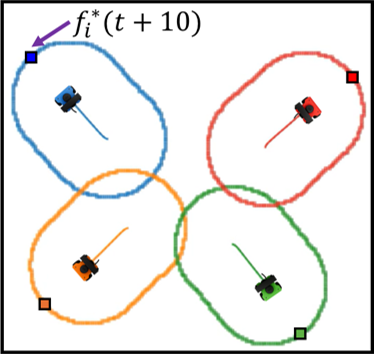}\label{fig:frontier_2}}%
    \vspace{-3pt}
    \caption{Example of the frontier selection process for exploration.}
    \label{fig:frontier_caption}
    \vspace{-18pt}
\end{figure}

Exploration is achieved via a frontier-based approach leveraging first-order particle propagation. Traditional techniques rely on communication to decide what actions to perform. In contrast, here each robot is predicting the expected state of the other teammates to decide where to go next.

To start, an $i^\text{th}$ robot will update its and the expected map of every other robot $j$ at the state described by the $j^\text{th}$ belief particle. Using this belief map, the $i^\text{th}$ robot can evaluate its frontier set $\mathcal F_i$ defined as the set of all cells that are adjacent to unknown cells (i.e., outside of $i^\text{th}$ field of view and not visited by $i$ yet) and pick a frontier $f_i^*$ to aim for, chosen among all $f_i\in\mathcal F_i$. This is illustrated in Fig. \ref{fig:frontier_caption}. To enforce a uniform exploration, in this work we decided to leverage a homogeneous frontier selection to track the $f_i^*$ that is in the direction most opposite to the other robots positions:
\begin{equation}
{f}_{i,\texttt{e}}^* \;=\;\underset{{f_i}\in\mathcal{F}_i}{\argmin}\;
\frac{\bigl(\bm{x}(f_i)-\bm{x}_i\bigr)^\top }
     {\lVert \bm{x}(f_i)-\bm{x}_i\rVert\,} \hat{\bm{d}}_i\,,
     \label{eq:frontier_selection}
\end{equation}
\begin{equation}
\bm{d}_i \;=\; \frac{1}{\,n_r-1\,}
\sum_{\substack{j=1 \\ j \neq i}}^{n_r}
\bigl(\bm{x}_j - \bm{x}_i\bigr)
\;\Longrightarrow\;
\hat{\bm{d}}_i \;=\; \frac{\bm{d}_i}{\lVert \bm{d}_i\rVert},
\end{equation}
where $\bm{d}_i$ is the mean displacement vector from robot $i$ to the estimated positions of all other robots, $||\bm{d}_i||$ is the length of $\bm{d}_i$, and $\hat{\bm{d}}_i$ is the normalized vector. 
The position of the frontier cell is $\bm{x}(f_i)$, ${f_i} \in \mathcal{F}_i$.

$i$ propagates these particles over time by predicting the control input and the resulting list of frontiers that robot $j$ would possess, $\mathcal{F}_j$ if $j$ continues to behave according to $b_i^j$
This evolves the state over time according to:
\begin{equation}
\label{eq:propagation}
\tilde{\bm{x}}_i^j(t+\ell+1)
=
h\bigl(\tilde{\bm{x}}^j_i(t+\ell),\;\tilde{\bm{u}}_j(t+\ell)\bigr),
\quad
\ell=1,\dots,{H},
\end{equation}
where $h(\cdot)$ is assumed to be a known dynamical model function of robot $j$.

Empathy particles are formed by propagating the nested state beliefs over ${H}$ using the previously described process and are used by $i$ to understand how predictable it is acting. For instance, if $i$ must avoid an obstacle that is unknown to $j$, $i$ will be forced to deviate from the location held within $^i{\mathcal{P}}_i^{j}(t)$. By creating $^i{\mathcal{P}}_i^{j}(t)$, robot $i$ can control back to the location that $i$ believes $j$ believes $i$ should be. 

\subsubsection{Repulsive Frontier} In addition to forming belief and empathy particles, a robot must recognize when its beliefs are incorrect and update its predictions. To do so, it compares its sensor readings with their predicted values to determine when an observation is unexpected. Such unexpected readings may occur naturally during exploration or be deliberately induced by peers. We formalize unexpectedness via a residual measure:
\begin{equation}
\label{eq:residual}
    \lambda_{i,j}(t) = \lVert \tilde{\bm{x}}_i^j(t) - \bm{x}_i^{j}(t) \rVert ,
\end{equation}
where $\bm{x}_i^{j}(t)$ is the observed state of $j$. If $\lambda_{i,j}(t)$ exceeds a predefined threshold, 
$i$ recalculates its frontier selection by selecting $f_i^*$ as the frontier most opposite of $\bm{x}_j$ using:
\begin{equation}
\tilde{f}_{j,\bar{\texttt{e}}}^* \;=\; \underset{\tilde{f_j}\in\tilde{\mathcal{F}}_i^j}{\argmin}\;
\hat{\bm{d}}(\bm{x}_i,\bm{x}(f_j))^\top\hat{\bm{d}}(\bm{x}_i,\bm{x}_j).
\label{eq:most_opposite_cell}
\end{equation}

$i$ then propagates belief and empathy particles assuming that $b_j = \bar{\texttt{e}}$ over $H$ as in \eqref{eq:propagation}.

\subsection{Bayesian Epistemic State Updating}

During frontier exploration, a robot can experience a variety of events $\bm{\zeta}$ that provide evidence about whether a robot has discovered a task. In collaborative robot operations, this list of events can be narrowed to the following: (i) {\em unexpected observation of another robot}, (ii) {\em prediction that another robot has discovered a task}, (iii) {\em expected observation of another robot}, (iv) {\em failing to observe another robot when expected to}, (v) {\em successfully intercepting another robot}, (vi) {\em failing to intercept another robot}, and (vii) {\em task discovery}. For example, if the $j^\text{th}$ robot unexpectedly detects another robot, $i$, that has deviated from its predicted path, $\bm{x}_i(t) \neq \tilde{\bm{x}}_i(t)$, $j$ will reason that $i$ probably possesses knowledge not held by $j$, increasing the likelihood that $i$ knows the location of the task since $j$ has not yet discovered it. In this section, we explain how a robot $i$ updates its epistemic state hypothesis, $\bm{e}_i$ about every other robot, based on evidence collected during these events.

Each robot $i$ maintains a distribution of probabilities $P$ for every possible combination of epistemic states $\bm{e}_i^j \in E$ (e.g., for a 3 robot case $E_i=\{\bm{e}_i^1=[0,0,0],\bm{e}_i^2=[0,0,1],\ldots \bm{e}_i^8=[1,1,1] \}$). Each probability is initialized to a user-defined value, setting a prior that gets updated as events occur to obtain a posterior via a Bayesian update process. In this work, an event occurrence signifies the discovery of a task by a robot $k$ which will increase the probability of all the epistemic states that contain $e_k=1$. For ease, probabilities are increased (or decreased) using a fixed increment (or decrement) as follows.
Given $\bm{e}_i=[e_1, e_2, \ldots, e_k, \ldots, e_{n_r}]$:
\begin{equation}
    P(\bm{e}_i^j) = 
    \begin{cases}
      P(\bm{e}_{i}^j) \cdot P^+ \; & \text{if} \; e_k=1 \; \wedge \; \zeta_k=1 \;\;\; \forall \bm{e}_i^j \in E\\
      P(\bm{e}_{i}^j) \cdot P^- \; &\text{if} \; e_k=1 \; \wedge \; \zeta_k=0 \;\;\; \forall \bm{e}_i^j \in E
    \end{cases}      
    \label{eq:joint_prob}
\end{equation}
where $1>P^+>P^->0$ are the fixed increment and decrement values respectively and $\zeta_k$ is an indicator binary variable set to 1 if an event is detected associated with $k$. Similar to occupancy grid mapping, this approach will reinforce probabilities related to epistemic states in which $k$ is believed to have found the task.


Once an event has occurred and the probabilities have been set, we search among all epistemic state combinations for the most likely one. Here we compare the predicted map $\mathcal{M}_j$ -- associated with the states from the first order particle set $\mathcal P_i^j$ for each robot $j$ -- with the current map $\mathcal{M}_i$ discovered by $i$ and task location (if known) to assess if any robot $j$ has discovered a task. Formally, if $i$ has discovered a task ${g}$:
\begin{equation}
e_j=
    \begin{cases}
1 & \text{if}\; e_i=1 \; \wedge \mathcal{M}_j\cap \{\bm{x}_g\} \neq \emptyset  \; \forall j\in \mathcal{R}\\
0 & \text{otherwise}
    \end{cases}
\end{equation}
We then use \eqref{eq:joint_prob} to update the likelihood associated with each $\bm{e}_i^j \in E$.

Finally, robot $i$ updates its hypothesis of $\bm{e}_i$ by selecting the maximum-a-posteriori hypothesis from the set of all possible hypotheses, $E$:
\vspace{-10pt}
\begin{equation}
\bm{e}_i^* \;=\; \argmax_{\bm{e}_i\in E} \; P(\bm{e}_i^j).
\label{eq:map_epistemic}
\end{equation}

\subsection{Behavior Selection via Depth-Limited Tree Search}
\label{sec:behavior_selection}


Once an event is detected, the Bayesian update process outlined in the previous section leads to an updated $\bm{e}_i^*$. In turn, this epistemic state is used to decide what behavior $b\in\mathcal{B}$ to perform to complete the task in the shortest amount of time, i.e, to satisfy \eqref{eq:policy_selection}. 
This multi-step process involves: i) updating the predicted teammate's behavior state vector $\bm{b}_i$ via higher-order reasoning informed by robot $i$'s belief and empathy particles, and ii) using $\bm{b}_i$ to inform selection of the $b_i\in\mathcal{B}$ that minimizes \eqref{eq:missionT}. 

\subsubsection{Behavior State Formation via Higher-order Reasoning}
\label{sec:peer_reasoning}

Behavior selection starts with estimating peers’ behaviors. $\bm{e}_i^*$ indicates to robot $i$ whether any $j\in \mathcal{R}$ is aware of the task, but does not encode information about the time in which $j$ discovered it or $j$'s behavior afterward. 
Robot $i$ discerns the time that $j$ found the task by leveraging the historical pose estimates $\tilde{\bm{x}}_j(t)$ encoded in $\left\{\mathcal{P}_i^j(\tau)\right\}_{\tau=0}^{t}$.

Specifically, the estimated discovery time, $^g\tilde{t}^{j}_i$, is the first time that $i$ predicts $\bm{x}_g$ was within $j$'s sensing range, $S_j$:
\begin{equation}
\label{eq:time_to_goal}
^g\tilde{t}^{j}_i
=\min\Big\{\, \tau \in [0,t] \;\big|\; \lVert \tilde{\bm{x}}_i^j(t)-\bm{x}_{\mathrm{g}}\rVert \le S_j \Big\}.
\end{equation}
If the $i^{\text{th}}$ robot believes that $j$ has discovered the task, $i$ infers the behavior $j$ would select after the discovery, $\tilde{b}_i^j$, by performing a depth-limited behavior tree search initialized at $^g\tilde{t}^{j}_i$. 
This behavior tree is constructed using state information contained within $i$'s second- and third-order beliefs to allow $i$ to reason about how $j$ believes the mission will progress based on any behavior $j$ selected. Each branch of the behavior tree represents the resulting epistemic state and estimated total mission time, $\tilde{\mathcal{T}}_i^j(b)$ for all $b\in\mathcal{B}$. 

\begin{figure}[t]
\centering
\includegraphics[width=0.47\textwidth]{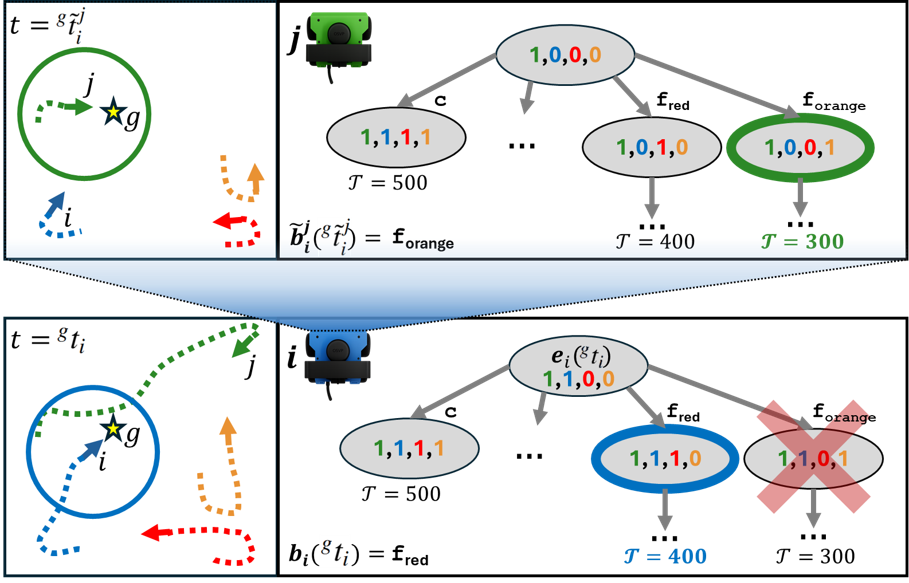}
\vspace{-7pt}
\caption{Illustration of the tree search-based behavior selection method.} 
\label{fig:behavior_tree}
\vspace{-15pt}
\end{figure}

Consider the example illustrated in Fig.  \ref{fig:behavior_tree} in which the {\em blue} robot discovers the task at $^gt_i$. {\em Blue} reasons that {\em green} has also seen the task. Using its second- and third-order empathy particles, {\em blue} determines that {\em green} would have likely selected the behavior, $\texttt{f}_{\text{orange}}$ after task discovery. {\em Blue} then prunes this branch from its behavior tree and selects the remaining option that will minimize $\mathcal{T}$ (i.e. $\texttt{f}_{\text{red}}$). 

To estimate $\tilde{\mathcal{T}}_i^j(b)$, $i$ must calculate when $j$ believes that each $k\in \mathcal{R}$ will complete the task $\forall b\in \mathcal{B}$. We denote this higher-order belief as $^k\tilde{T}_i^j$, read as $j$'s belief of the time required for $k$ to complete the task as computed by $i$. In other words here $i$ is rewinding time after discovering that $j$ was at the same task to recreate $j$'s reasoning process given its knowledge at the time about all $k\in \mathcal{R}$. 


For $b \in \big\{\mathtt{c}, \mathtt{e}, \mathtt{e}^*\big\}$, $^k\tilde{T}_i^j$ is computed by casting the problem as a forward reachability analysis with
\begin{equation}
^k\tilde{T}_i^j=\min\{t\;|\;(\mathcal{S}_k,\; t\in[^g\tilde{t}^{j}_i,^g\tilde{t}^{j}_i+H])\cap \{\bm{x}_g\} \neq \emptyset\}
\end{equation}
where $\mathcal{S}$ is the reachable set computed by integrating forward the dynamics of $k$ considering the sensing field of view area described by $k$ while moving \cite{esen_work}.

If $b = \mathtt{f}$, $^k\tilde{T}_i^j$ is estimated using a Dubins reachability analysis \cite{dubinsreachability} to estimate the time required for $j$ to reach the point that will lead to a successful fetch of $k$ and return to the task (more details about fetching is provided in Sec. \ref{sec:mppi}). 


With $^k\tilde{T}_i^j$, the maximum mission time is expressed as:
\begin{equation}
\tilde{\mathcal{T}}_i^j(b)= \left[\max_{k\in \mathcal{R}_g}{^k\tilde{T}_i^j},\right]
\end{equation}
and the behavior of $j$ is obtained by:
\begin{equation}
\label{eq:peer_behavior}
\tilde{b}_i^j
= \argmin_{b \in \mathcal{B}} 
 \tilde{\mathcal{T}}_i^j(b).
\end{equation}

\subsubsection{Behavior Selection}
Robot $i$, that has discovered the task, now has a belief of who else discovered the task ($\bm{e}$) and what behavior they are now performing ($\bm{b}$). With this information, $i$ can select its optimal behavior using the same behavior tree search process above.
$i$ can prune the behavior tree to remove behaviors that it believes have already been performed (i.e., not fetching robot $k$ because $i$ has predicted that $j$ likely decided to fetch $k$ when $j$ previously found the task). Given this pruned tree, $i$ can then select the best behavior which minimizes cost according to: 
\begin{equation}
\label{eq:self_behavior}
b_i
= \argmin_{b \in \mathcal{B}} 
\mathcal{T}_i.
\end{equation}

\subsection{Interception via Temporally Aware MPPI}
\label{sec:mppi}
Upon discovering a task, if robot $i$ selects {\em fetch} $\texttt{f}$, it uses $\mathcal{P}_i^j(t)$ to plan an intercept that redirects $j$’s frontier choice toward faster discovery of the task. This is possible because $i$ knows that an unexpected appearance in $j$’s field of view raises the residual, prompting $j$ to switch to the modified frontier approach $\bar{\texttt{e}}$ in \eqref{eq:most_opposite_cell}. Executing a fetch behavior $\texttt{f}$ involves choosing where and when to meet $j$, generating a trajectory to reach the intercept point and assessing success by updating the predicted trajectory of $j$.
We implement this with a modified MPPI controller, \cite{mppi}, which efficiently samples diverse candidates and returns a stochastically optimal intercept path. As shown in Fig.~\ref{fig:Mppi_baseline}, a non-optimal intercept can be counter-productive to {\em blue}’s goal of pushing {\em green}’s search toward the task; with our MPPI (Fig.~\ref{fig:Mppi_motivation}), {\em blue} avoids unintended rendezvous and arrives at the right place and time. The next section details intercept-point selection, trajectory generation, and success evaluation.
\subsubsection{Finding the ideal intercept point}
\label{sec:find_ideal_intercept}
To execute a fetch behavior, $i$ first identifies $\bm{x}^*_{\mathrm{int}}$, the earliest reachable position that causes $j$ to detect $i$ and select repulsive frontiers that drive $j$ toward the task. The set of candidate intercept points, $\mathcal{I} = \{\bm{x}_{\mathrm{int}}(\tau)\bigr\}_{\tau=t}^{t+H}$, is generated over a finite horizon $H$, and the earliest candidate that is deemed reachable by $i$ via Dubins reachability analysis is selected.
Each $\bm{x}_{\mathrm{int}}(\tau)$ is constructed as the point just inside $j$’s sensing range, $S_j$, on the side opposite from $\bm{x}_g$ at $t=\tau$:
\begin{equation}
\bm{x}_{\mathrm{int}}(\tau) =
\tilde{\bm{x}}_j(\tau) + (S_j-\sigma)\,\hat{\bm{d}}(\bm{x}_g,\tilde{\bm{x}}_j(\tau)),
\label{eq:reachable_set}
\end{equation}
where $\sigma$ is a small offset, placing $i$ within $j$'s sensor range. We compute the time required for robot $i$ to reach each candidate intercept point $\bm{x}_{\mathrm{int}}(\tau)\in \mathcal{I}$ using Dubins time~\cite{dubinsreachability}, which combines the straight line travel time and heading alignment time to $\bm{x}_{\mathrm{int}}$:
\begin{equation}
\label{eq:dubins}
\begin{aligned}
D\bigl(\bm{x}_{\mathrm{int}}(\tau), \bm{x}_i(t)\bigr)
&= \frac{\bigl\|\bm{x}_{\mathrm{int}}(\tau)-\bm{x}_i(t)\bigr\|}{v_{\max}}
\\
&+ \frac{1}{\omega_{\max}}\,
   \arccos\Bigl(
      \hat{\bm{\theta}}_i(t)\cdot
      \hat{\bm{d}}\bigl(\bm{x}_i(t), \bm{x}_{\mathrm{int}}(\tau)\bigr)
   \Bigr).
\end{aligned}
\end{equation}
 where $\hat{\bm{\theta}}_i(t) = \left[\cos(\theta_i(t)), \sin(\theta_i(t))\right]$ is the unit orientation vector, $v_{\max}$ is robot $i$'s maximum linear velocity, and $\omega_{\max}$ is the maximum angular velocity. 

The earliest feasible candidate is selected by minimizing the index $\tau$ subject to reachability:
\begin{equation}
\tau^*
=
\argmin_{\tau\in\{t,\dots,t+H\}}
\;(\tau - t)
\quad
\text{s.t.}\quad
D\;\le\;\alpha\,(\tau - t).
\label{eq:feasible_min}
\end{equation}

Because $D$ assumes a direct route at maximum speed, we introduce a slack factor $\alpha \le 1$ to account for longer, non-direct trajectories that may be generated by the MPPI.

The optimal intercept point is then set as $\bm{x}^*_{\mathrm{int}}=\bm{x}_{\mathrm{int}}(\tau^*)$, which we send to the MPPI for trajectory generation. 


\subsubsection{MPPI Trajectory Selection}
\label{sec:mppi_traj}

\begin{figure}[t]
    \centering
    \subfigure[Baseline direct trajectory]{
    \includegraphics[width=0.232\textwidth]{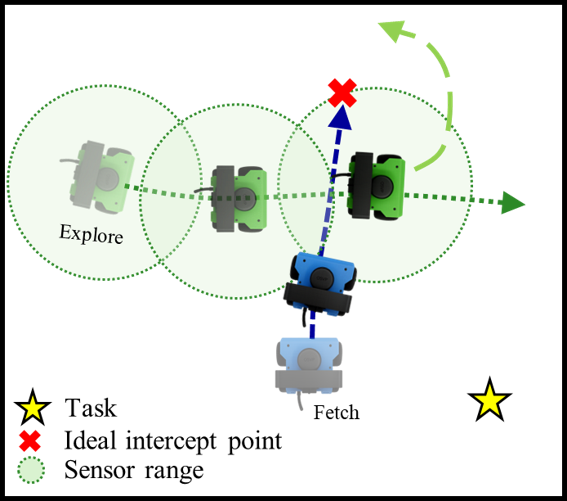}\label{fig:Mppi_baseline}}
    \subfigure[MPPI generated trajectory]{
    \includegraphics[width=0.232\textwidth]{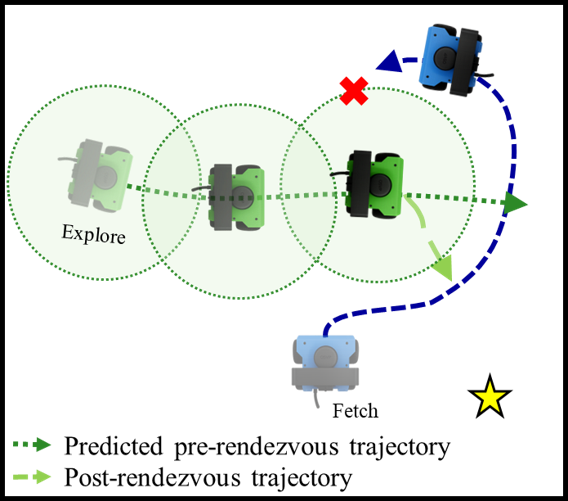}\label{fig:Mppi_motivation}}
    
    \vspace{-7pt}
    \caption{Illustration of the fetching behavior: (a) a naive approach that leads to early rendezvous causes a counter-productive reaction by {\em green}; (b) with MPPI, rendezvous is avoided until the optimal time.}
    \label{fig:mppi_motivation_combined}
    \vspace{-15pt}
\end{figure}
Given $\bm{x}^*_{\mathrm{int}}$, we use a modified MPPI to iteratively generate a stochastically optimal trajectory $\bm{u}^*=\{\bm{u}_1,\dots, \bm{u}_{H_u}\}$ over horizon $H_u$. Following \cite{mppi}, at each iteration we create $K$ candidate control sequences $\{\bm{u}_k\}_{k=1}^K$, and use the robot's dynamic model to obtain candidate trajectories $\{\bm{r}_k(h)\}_{k=1}^K$ with $h=\{0,\dots , H_u\}$ being the discrete time index. We evaluate each candidate trajectory with a three-part cost function that penalizes misalignment between $i$ and $j$, distance from $\bm{x}_{int}^*$, and safety:
\begin{equation}
\begin{aligned}
J(\bm{r}_k) &= \sum_{h=0}^{H_u} m(\bm{r}_k, h)\Phi(\bm{r}_k, h){} +
   \bigl\lVert \bm{r}_{k}(h) - \bm{x}^*_{\mathrm{int}}\bigr\rVert^2
 +  c_{\mathrm{s}}\bigl(\bm{r}_{k}(h), \mathcal{M}\bigr)     .
\end{aligned}
\end{equation}
The first term encourages an intercept geometry in which robot $i$ positions itself on the side of $j$ opposite the task $\bm{x}_g$ (Fig.~\ref{fig:Mppi_motivation}). It activates only when $i$ is within $j$’s sensor range $S_j$ and then penalizes same-side alignment with respect to the task. Concretely, $m=\eta$ if $||\tilde{\bm{x}}_j(h)-\bm{r}_k(h)||\le S_j$, otherwise $0$, where $\eta$ is a constant gain applied to the alignment cost. $\Phi$ is defined as:
\begin{equation}
\label{eq:K_and_Phi}
\Phi(\bm{x}_k, h)
=
\begin{cases}
(\Psi(\bm{r}_k, h)+2)^2, & \text{if}\;\;\Psi(\bm{r}_k,h) > \gamma,\\[3pt]
0, & \text{else},
\end{cases}
\end{equation}
where $\Psi(\cdot)=\hat{\bm{d}}(\tilde{\bm{x}}_j(h), \bm{x}_g)^\top\hat{\bm{d}}(\tilde{\bm{x}}_j(h), \bm{r}_k(h))$, measures the dot product between the vector from $\tilde{\bm{x}}_j(h)$ to $\bm{x}_g$ and the vector from $\tilde{\bm{x}}_j(h)$ to $\bm{r}_k(h)$. $\gamma$ is a user-defined threshold.

The second term in $J(\bm{r}_k)$ promotes liveness by penalizing deviation from the intercept point, while the third term penalizes unsafe proximity to known obstacles via $c_{\mathrm{s}}(\bm{r}_k(h),\mathcal{M})$ when the separation between $\bm{r}_k(h)$ and any $\bm{x}_{\mathrm{obs}}\in\mathcal{M}$ falls below a safety margin.


After evaluating $J_k$ for all $K$ candidate trajectories, the optimal sample index is selected as:
\begin{equation}
    k^* = \argmin_{k} J^{(k)}.
\end{equation}
The corresponding control sequence is $\bm{u}_{k^*}$.

After executing the MPPI trajectory $\bm{x}_i^*$, robot $i$ enters $j$’s field of view, prompting $j$ to evaluate the residual \eqref{eq:residual}; if $\lambda_{i,j}(t)>\delta$, $j$ replans with the modified exploration policy $\bar{\texttt{e}}$ and selects a new frontier. Anticipating this, $i$ propagates an updated belief particle to predict $j$’s new trajectory with $b_j = \bar{\texttt{e}}$. The intercept is declared successful if:  
\begin{equation}
^gt_{i}^{j}(t)
\;\le\;\kappa\,D(\tilde{\bm{x}}_j(t),\;
\bm{x}_{\mathrm{g}}),
\end{equation}
where $D$ is the Dubins time to task for $j$, as calculated in \eqref{eq:dubins} and $\kappa$ is a safety factor that allows indirect routes. If unsuccessful, $i$ selects a new $\bm{x}^{*}_{\mathrm{int}}$ and retries. If $i$ determines the intercept was successful, $i$ updates its epistemic state and selects a new behavior. 
If the intercept was unsuccessful, robot $i$ repeats the MPPI process by selecting a new $\bm{x}^*_{\mathrm{int}}$ and attempts another intercept. 

\section{Simulations}\label{sec:simulations}



Our approach was evaluated in a simulated $150\text{m}^2$ environment. Sensing was performed using a simple lidar sensor model with a range of 20m. Robots were modeled as differential-drive with nominal speed of 1.0m/s, and the ability to increase speed up to 1.25m/s when required for fetching or returning to a robot's second-order empathy particle. This formulation reflects a common practice in long-duration robotic operations, where a nominal endurance speed is maintained for efficiency, but higher speeds can be used for short bursts when needed \cite{endurance}. 
Simulations were performed with teams of 2, 3, and 4 robots over 100 randomized trials per team size, sampling task locations and robots’ initial poses uniformly within the environment. For each task, all robots were needed to complete the task, i.e., $\mathcal{R}_g = \mathcal{R}$ and task completion was defined as all team members reaching within 1.0m of the task. We considered both: (i) empty environments and (ii) cluttered environments containing five square obstacles placed at random locations. 

As a comparison baseline, we implemented a first-order reasoning policy where each robot predicts peers’ positions using only first-order beliefs biasing its exploration toward regions opposite to those peers, as explained in Sec. \ref{sec:epistemic_belief}. 
\begin{figure}[t]
\centering
\includegraphics[width=0.47\textwidth]{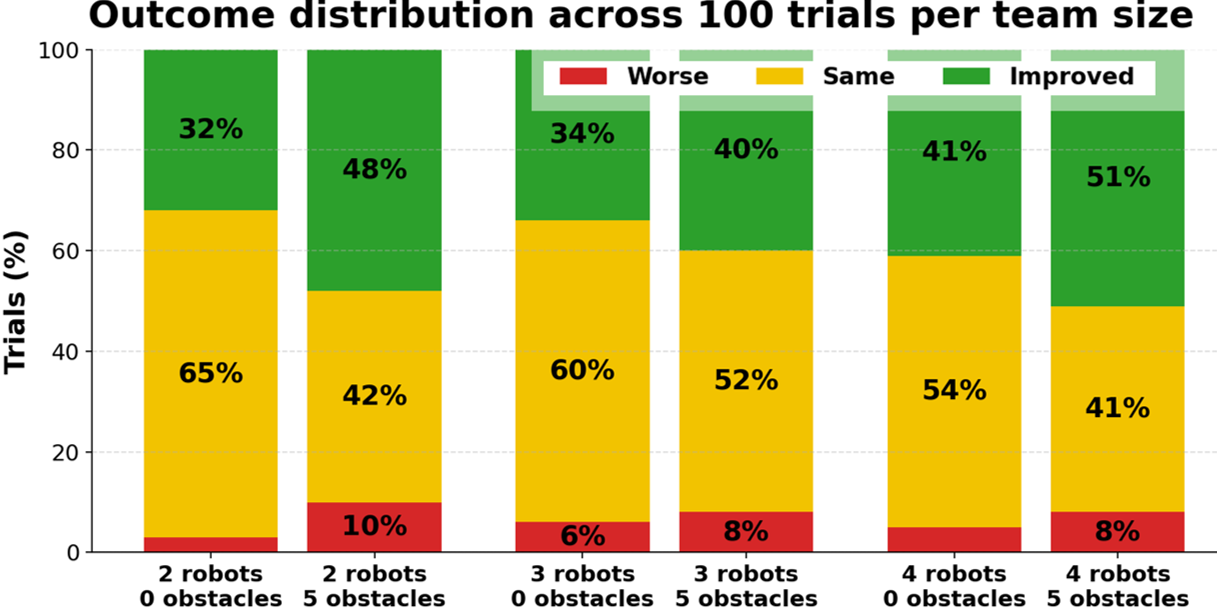}
\vspace{-7pt}
\caption{Outcome distribution across the 6 simulated scenarios. Our approach is deemed successful if the time to completion is the same (yellow) or improved (green) compared to the baseline.}
\label{fig:sim_stacked_bar}
\vspace{-15pt}
\end{figure}

\begin{figure}[t]
\centering
\includegraphics[width=0.47\textwidth]{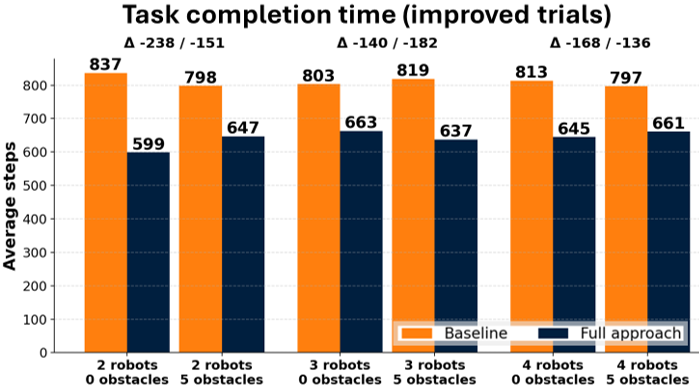}
\vspace{-5pt}
\caption{A comparison of the task completion time for our approach versus the baseline when a decision to fetch was made by a robot within the MRS (improved trials from Fig.~\ref{fig:sim_stacked_bar}).}
\label{fig:sim_comparison}
\vspace{-10pt}
\end{figure}

Across $600$ trials ($100$ per setting), our approach matched or outperformed the baseline in $93.3\%$ of cases, as shown in Fig.~\ref{fig:sim_stacked_bar}. A ``match'' occurs when a robot that finds the task predicts that all teammates will arrive at the task sooner than a fetch maneuver could be accomplished. The time-optimal decision is therefore to abstain from fetching, as fetching would not reduce completion time.
When our policy included a fetch behavior, it reduced completion time by an average of $169$s (20.8\%), shown in Fig.~\ref{fig:sim_comparison}.

\begin{figure}[t]
    \centering
    \subfigure[$t=14$s]{%
        \includegraphics[width=0.32\columnwidth]{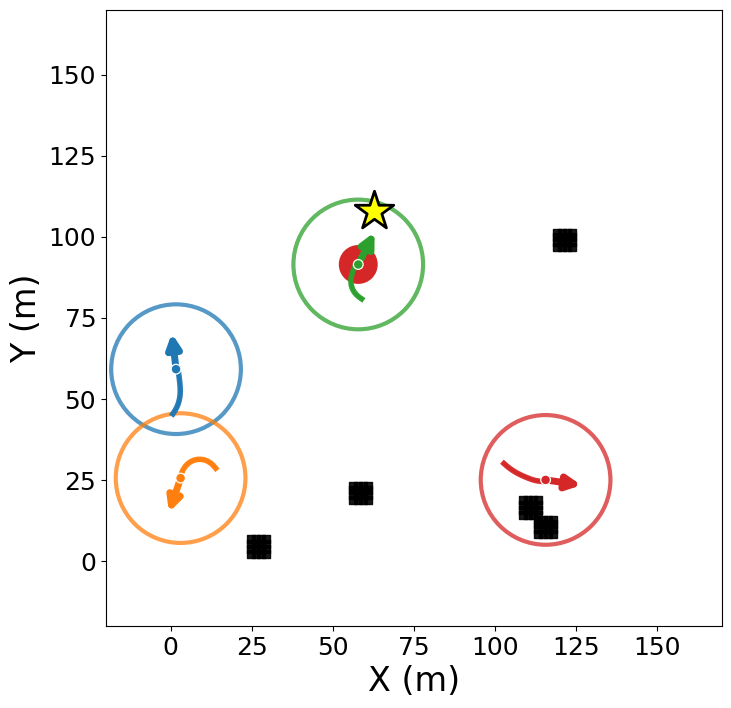}%
        \label{fig:sim1}%
    }%
    \subfigure[$t=147$s]{%
        \includegraphics[width=0.32\columnwidth]{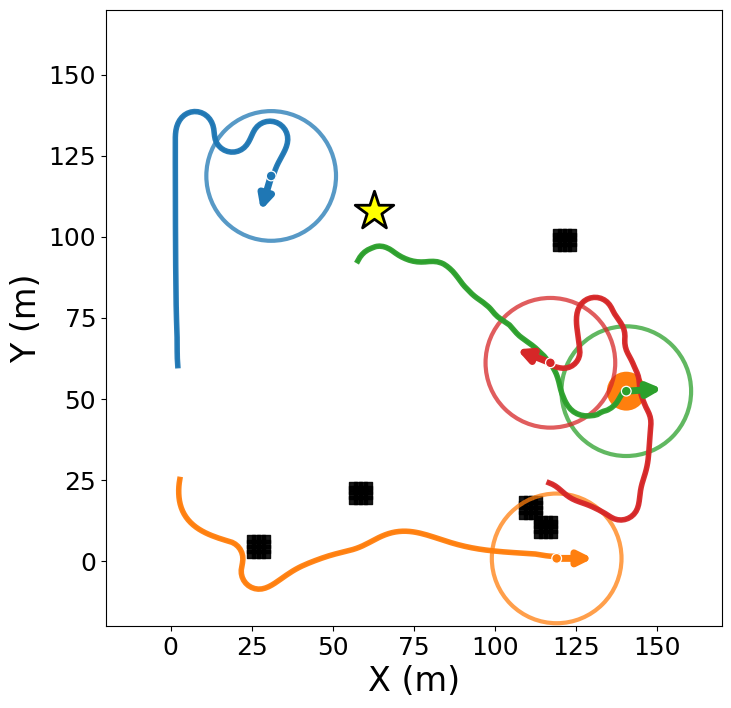}%
        \label{fig:sim2}%
    }   
    \subfigure[$t=205$s]{%
        \includegraphics[width=0.32\columnwidth]{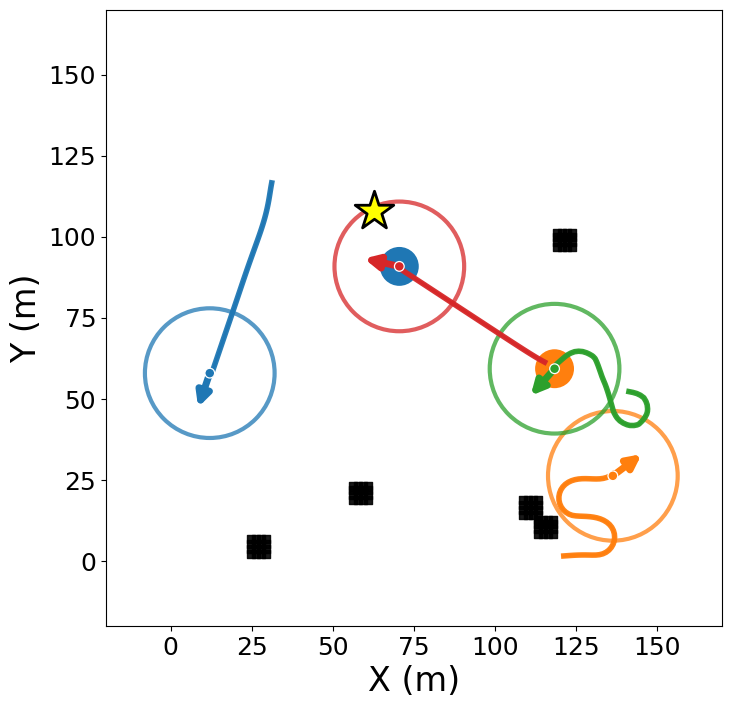}%
        \label{fig:sim3}%
    }   
    \subfigure[$t=375$s]{%
        \includegraphics[width=0.32\columnwidth]{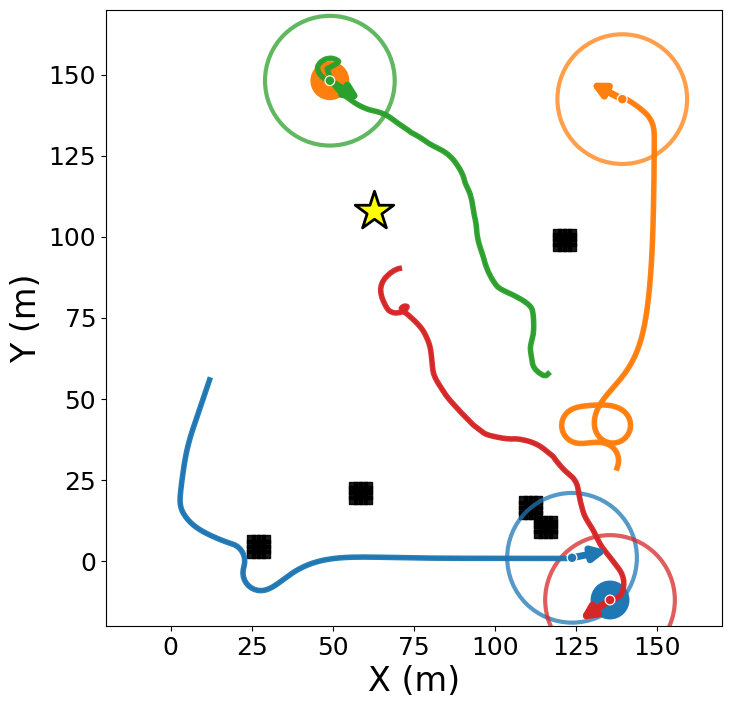}%
        \label{fig:sim4}%
    }    
    \subfigure[$t=431$s]{%
        \includegraphics[width=0.32\columnwidth]{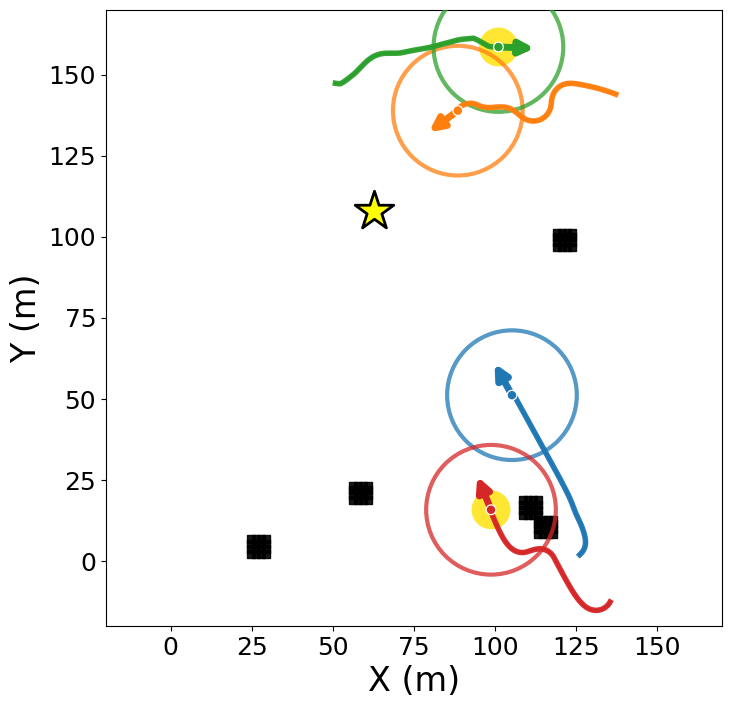}%
        \label{fig:sim5}%
    }%
    \subfigure[$t=529$s]{%
        \includegraphics[width=0.32\columnwidth]{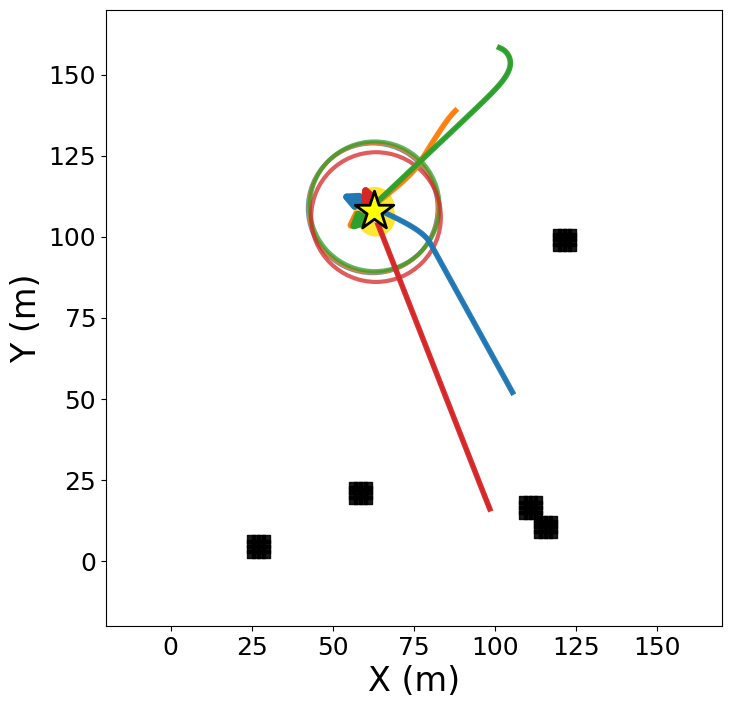}%
        \label{fig:sim6}%
    }
    \vspace{-7pt}
    \caption{Simulation snapshots of a 4-robot case in a 5-obstacle environment. Colored rings denote sensing range; circles encode behavior as: yellow = complete task; robot’s color = fetching that robot; none = explore.}
    \label{fig:sim_results}
    \vspace{-20pt}
\end{figure}

These improvements arise from the higher-order reasoning capability of our framework. An illustrative example is shown in Fig.~\ref{fig:sim_results}. In this scenario, the {\em green} robot detects the task in Fig. \ref{fig:sim1}, recognizes that it is the only robot aware of the task, and identifies that fetching {\em red} yields the greatest time reduction. As shown in \ref{fig:sim2}, {\em green} uses the MPPI, to position itself to intercept {\em red} and, upon confirming a successful interception, {\em green} repeats this process. The next most beneficial action is to fetch {\em orange}. While {\em green} fetches {\em orange}, {\em red} discovers the task, as shown in Fig. \ref{fig:sim3}. Leveraging its second- and third-order beliefs about {\em green}, {\em red} predicts that {\em green} would have fetched both {\em red} and {\em orange} and reasons that fetching {\em blue} is the most beneficial behavior remaining. {\em Red} completes the intercept in Fig. \ref{fig:sim4} while {\em green} completes its intercept in Fig. \ref{fig:sim5}. Through this sequence of higher-order-based reasoning, all robots converge on the task in $529$s: $333$s faster than the baseline, shown in the supplementary material. 

Figure~\ref{fig:sim2} illustrates additional behaviors arising from higher-order reasoning. When the {\em orange} robot encounters an obstacle, it adjusts its trajectory while maintaining a second-order belief of its perceived location by teammates, allowing it to return gradually to the position consistent with its second-order empathy particle. This adjustment enables the {\em green} robot to successfully intercept {\em orange} later in the simulation at $t = 431$s. 
Such behaviors demonstrate how higher-order epistemic reasoning facilitates coordination, even in unknown and cluttered environments.
These results show that our long-horizon estimation–planning–coordination framework based on higher-order reasoning significantly reduces task completion time relative to first-order baselines. We next present experiments with multiple ground robots to validate the feasibility of our approach in physical systems.


\section{Experiments}\label{sec:experiments}
\begin{figure*}[t]
    \vspace{-10pt}
    \centering
    \subfigure[snapshot at $t = 4$s]{%
        {\includegraphics[width=0.24\textwidth]{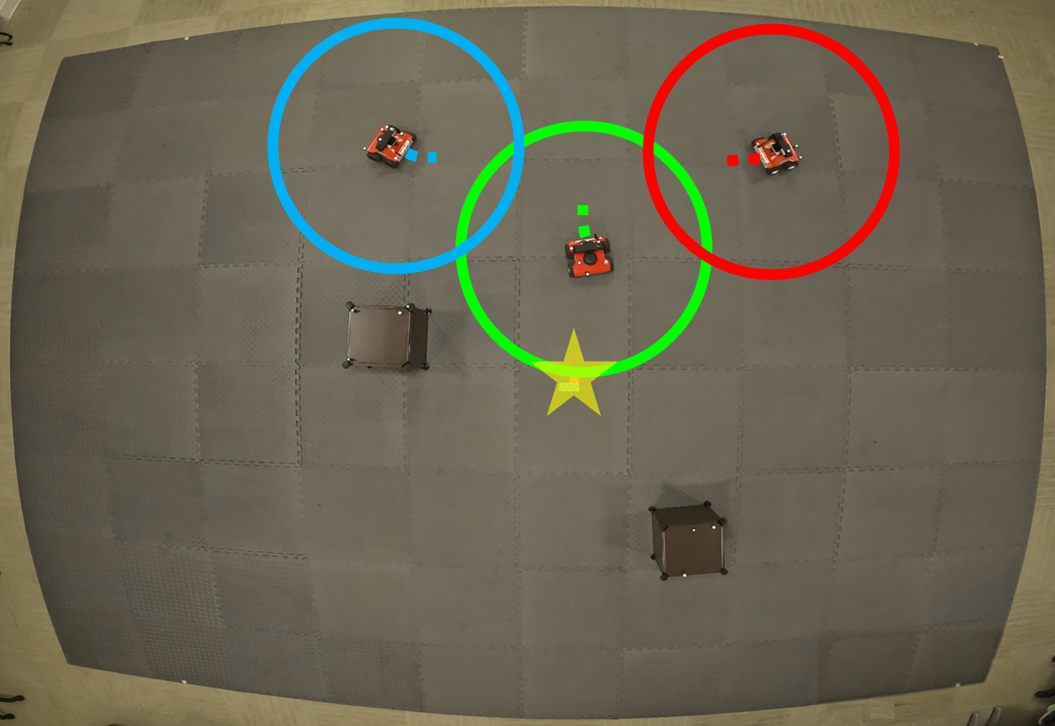}}%
        \label{fig:exp1}%
    }%
    \hfill
    \subfigure[snapshot at $t = 35$s]{%
        {\includegraphics[width=0.24\textwidth]{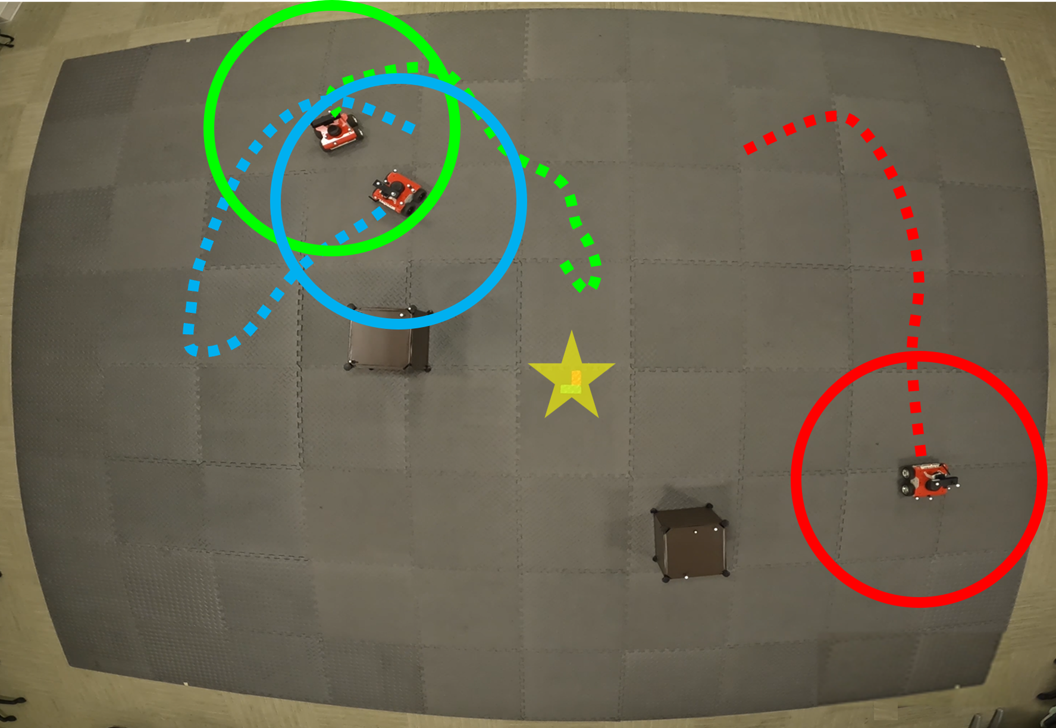}}%
        \label{fig:exp2}%
    }%
    \hfill
    \subfigure[snapshot at $t = 90$s]{%
        {\includegraphics[width=0.24\textwidth]{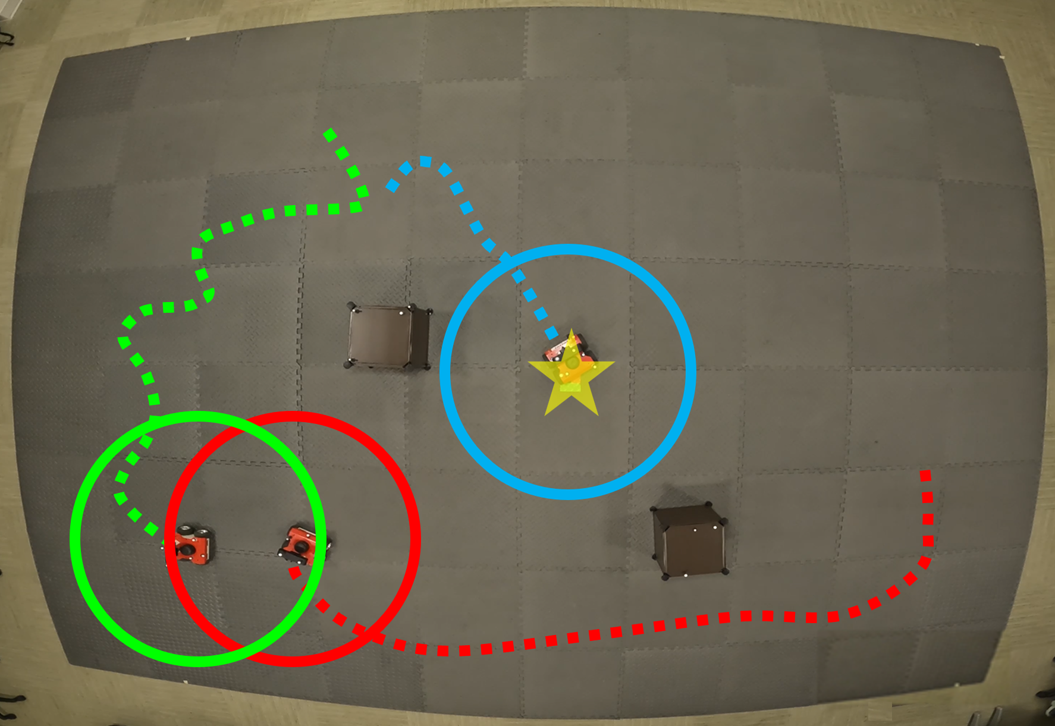}}%
        \label{fig:exp3}%
    }%
    \hfill
    \subfigure[snapshot at $t = 115$s]{%
        {\includegraphics[width=0.24\textwidth]{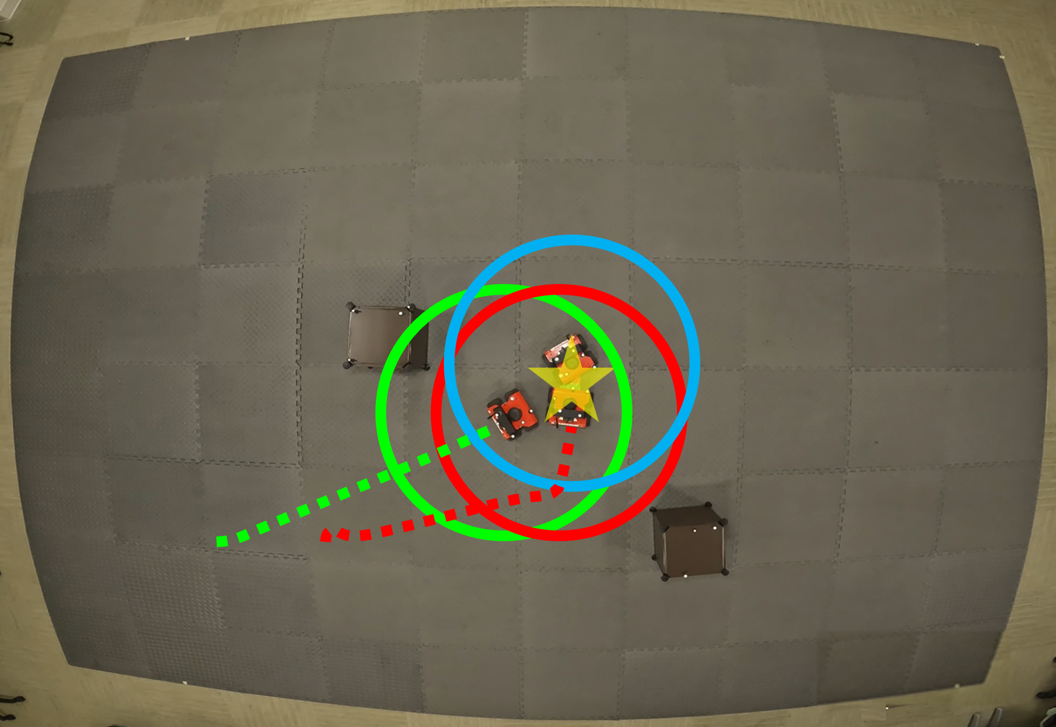}}%
        \label{fig:exp4}%
    }\\[-0.85em]
    \subfigure[Vicon data at $t = 4$s]{%
        {\includegraphics[width=0.24\textwidth]{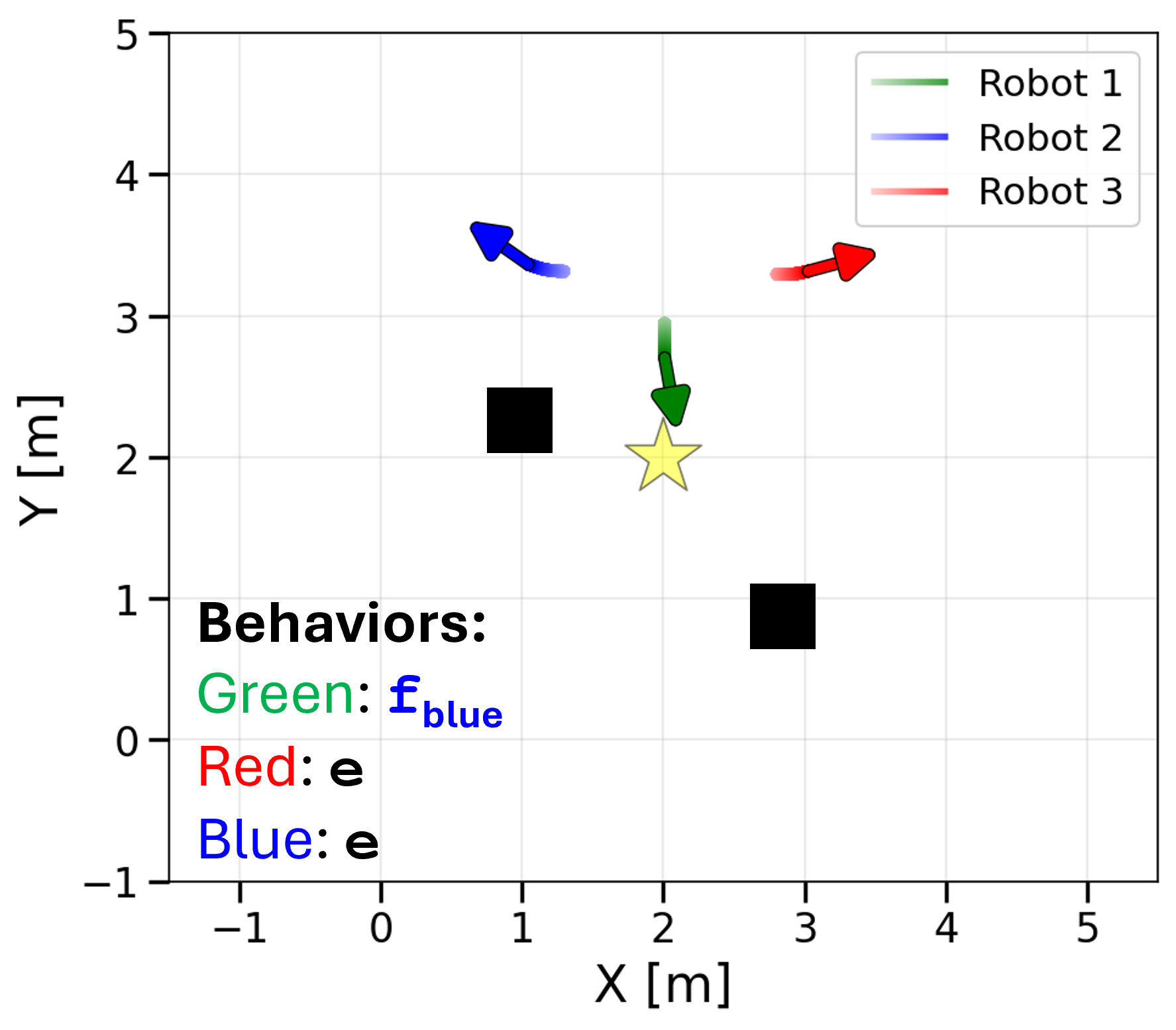}}%
        \label{fig:exp_data1}%
    }%
    \vspace{-2pt}
    \hfill
    \subfigure[Vicon data at $t = 35$s]{%
        {\includegraphics[width=0.24\textwidth]{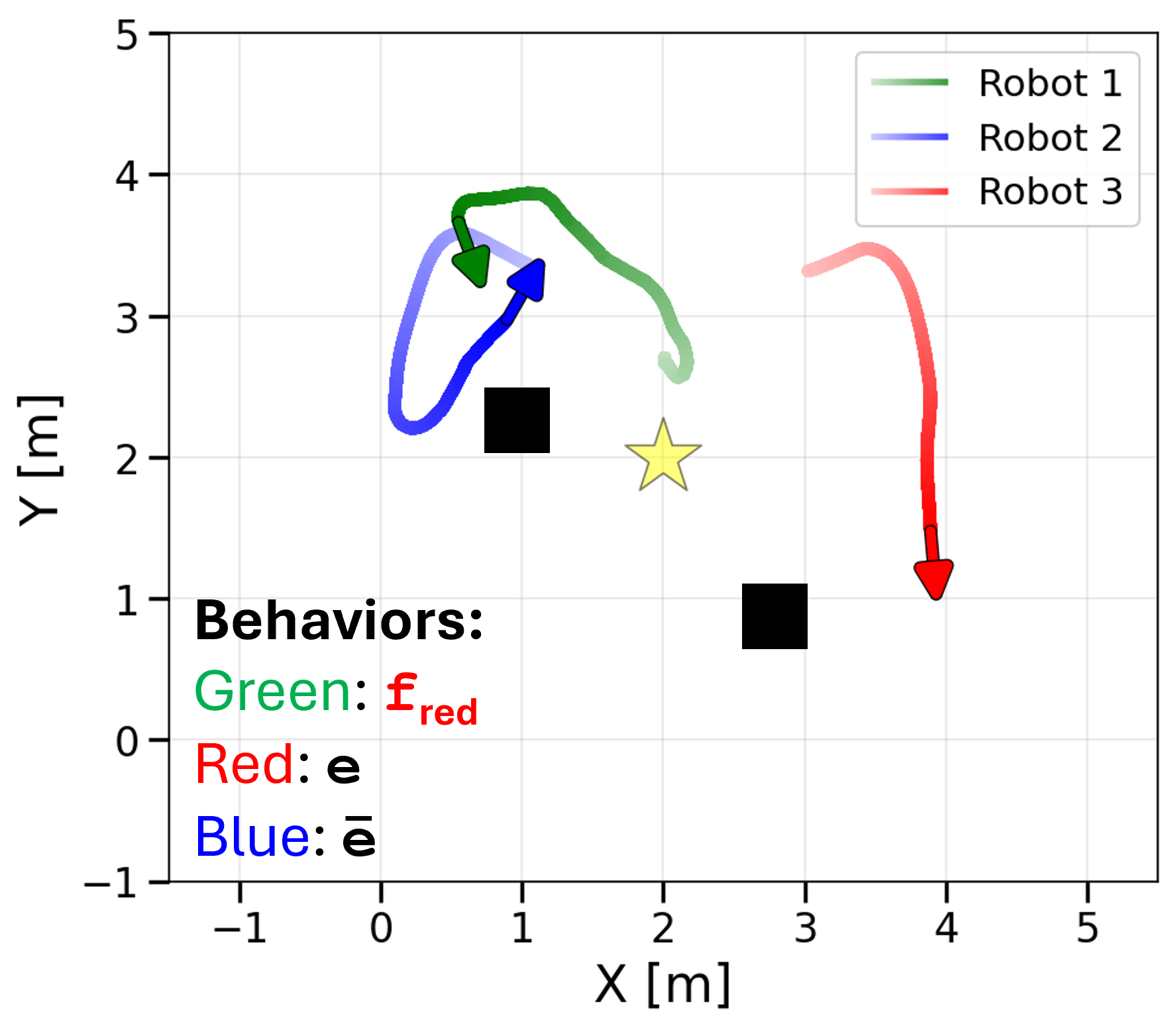}}%
        \label{fig:exp_data2}%
    }%
    \vspace{-2pt}
    \hfill
    \subfigure[Vicon data at $t = 90$s]{%
        {\includegraphics[width=0.24\textwidth]{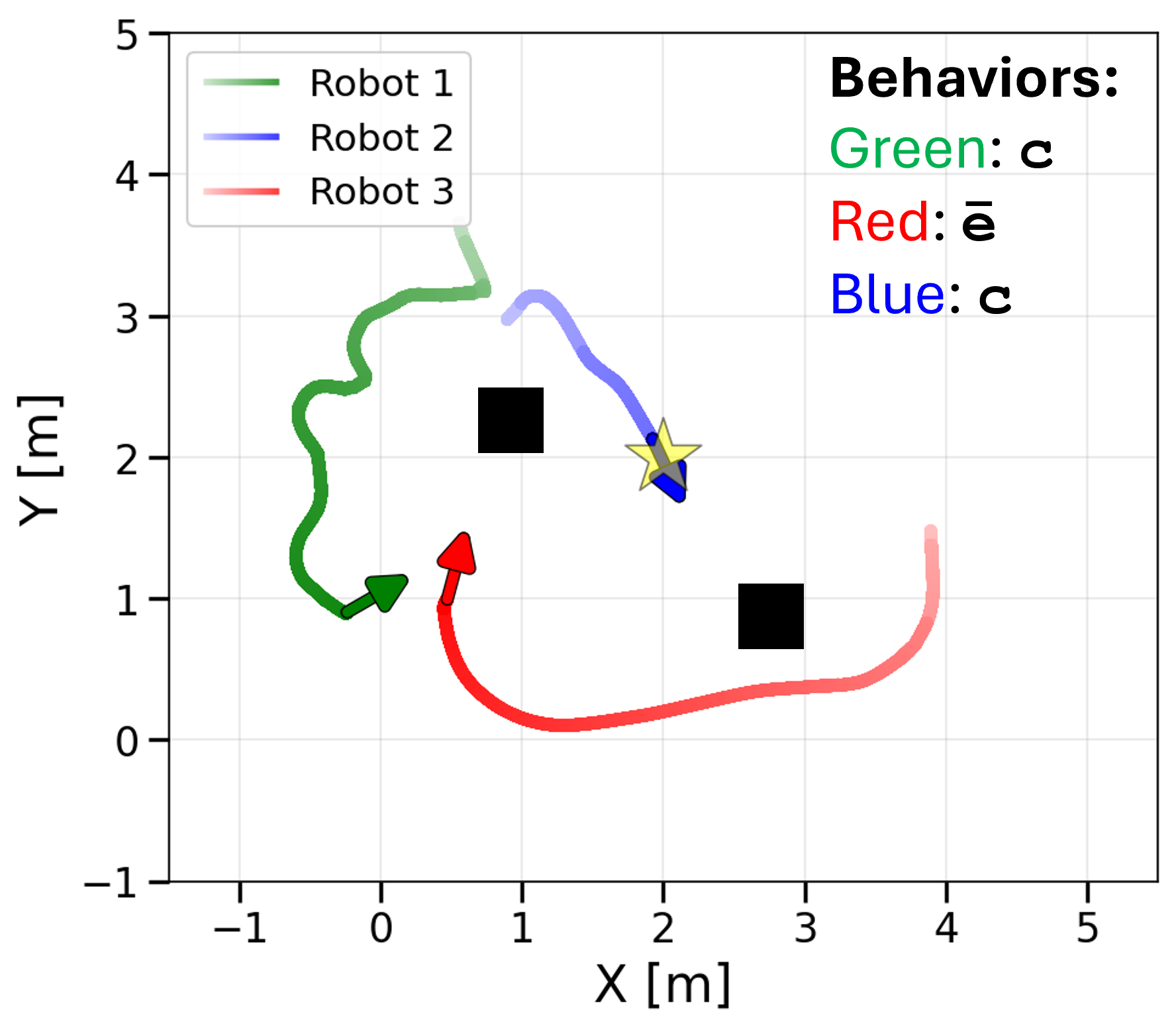}}%
        \label{fig:exp_data3}%
    }%
    \hfill
    \vspace{-2pt}
    \subfigure[Vicon data at $t = 115$s]{%
        {\includegraphics[width=0.24\textwidth]{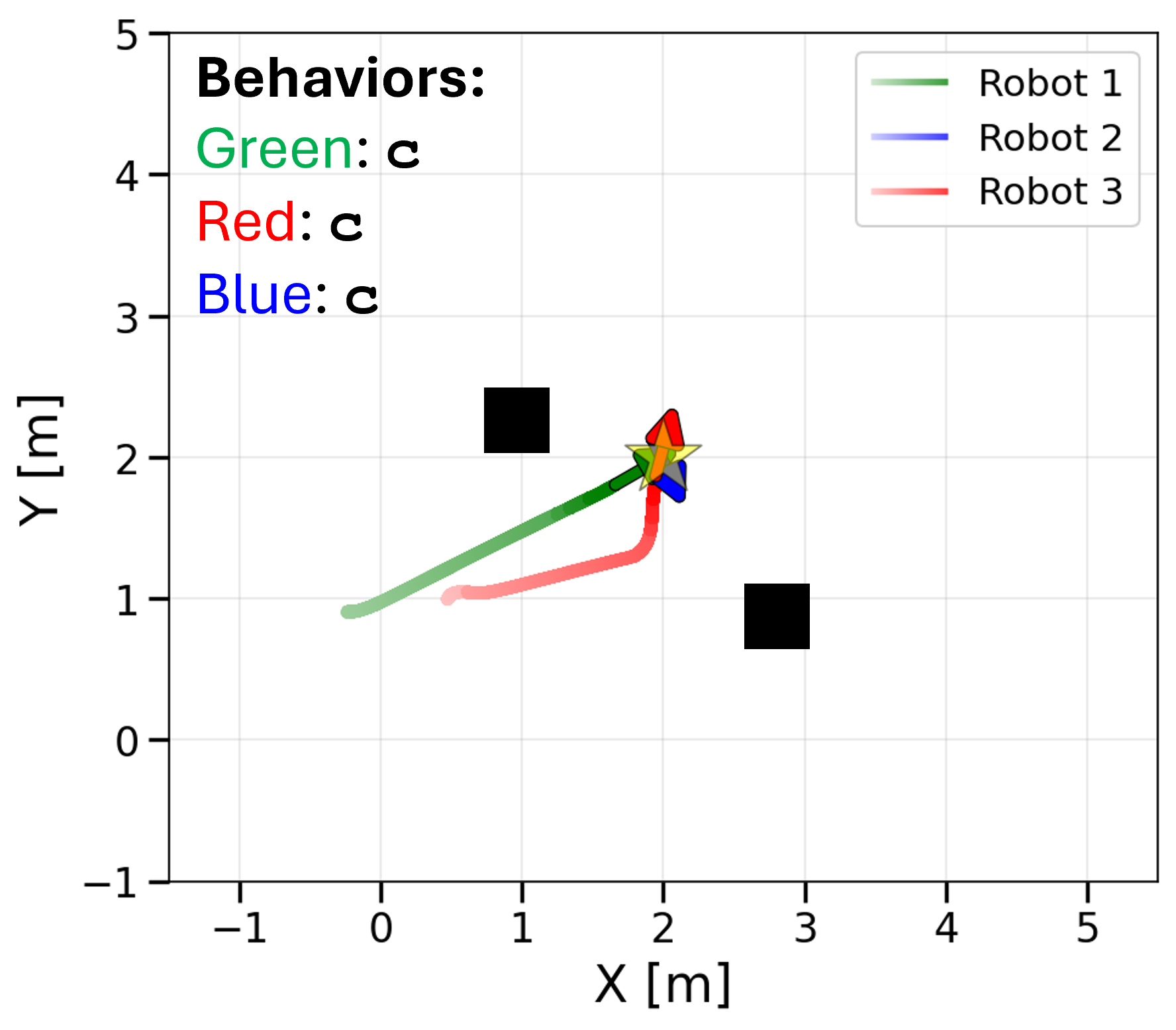}}%
        \label{fig:exp_data4}%
    }
    \vspace{-2pt}
    \caption{Experiment results for a 3 robots case.}
    \label{fig:experiment_results}
\vspace{-20pt}
\end{figure*}

Our approach was validated through laboratory experiments with teams of 2 and 3 robots placed at random locations within a $4\text{m}^2$ area. Because of the smaller operating region, the sensing range was decreased to $0.75 \,\text{m}$ and the nominal speed was set to $0.1 \,\text{m/s}$. As in the simulations, robots could increase their velocity up to $0.125 \,\text{m/s}$ during fetching or while returning to their second-order empathy particle after avoiding an obstacle. The task requires $\mathcal{R}_g = \mathcal{R}$. 

One example is shown in Fig. \ref{fig:experiment_results}, with the task placed in the center at $(2.0 \,\text{m}, 2.0 \,\text{m})$, and two $(0.3\text{m} \times 0.3\text{m}) $ obstacles placed within the environment.
As illustrated in Fig. \ref{fig:exp1}, the team of robots begins exploring the area, at which point the {\em green} robot locates the task and reasons that no other robots have detected it. As shown in Fig. \ref{fig:exp2}, {\em green} chooses to fetch {\em blue} using an MPPI. 
After evaluating the effectiveness of the intercept, the {\em green} robot determines that it succeeded fetching and selects a new behavior using the behavior tree process. This time, it finds that it can further minimize~\eqref{eq:missionT} by choosing to fetch {\em red}. 

While {\em green} is fetching {\em red}, {\em blue} discovers the task. Upon doing so, it applies higher-order reasoning to infer that {\em green} had already found the task and decided to fetch {\em blue}. {\em Blue} then reasons that this behavior was completed and that {\em green} would have chosen to fetch {\em red} as its next action. Based on this reasoning, {\em blue} ultimately selects complete task as its behavior, shown in Fig. \ref{fig:exp3}. 

The {\em green} robot successfully intercepts the {\em red} robot, shifts its search pattern toward the task, and enables mission completion, shown in Fig. \ref{fig:exp4}. This mission was completed in $t=115$s, a reduction of $84$s (57.8\%) compared to the baseline time of $t=199$s. Full experimental videos are provided in the supplemental material.
\section{Conclusions and Future Work}\label{sec:conclusion}
\vspace{-1pt}
This paper presented a decentralized framework for implicit multi-robot coordination in communicationless settings. Using dynamic epistemic logic, robots maintain belief and empathy particles to interpret unexpected events and reason about teammates; a higher-order reasoning-based behavior tree selects actions that minimize task completion time and an MPPI enables long-horizon planning under partial observability. Simulations and experiments show consistent reductions in task completion time versus first-order baselines, validating epistemic reasoning for resilient coordination. Although we focus on communicationless cases, this method naturally extends to limited-range communication systems, in which case the MPPI process can be simplified.


Future work aims to scale the system to larger and more complex environments with multiple tasks and to extend it to heterogeneous teams where robots must reason about the capabilities of other robots and task requirements. We also plan to investigate adversarial scenarios, where higher-order reasoning and epistemic logic may be leveraged not only for cooperation but also for countering deceptive or competitive behaviors and in human-robot scenarios where communication between human and robot is often not possible.
\vspace{-5pt}

\section{Acknowledgment}\label{sec:ack}
We gratefully acknowledge DARPA YFA award \#D25AC00371 and the Commonwealth Cyber Initiative for partial support of this work.



\bibliographystyle{IEEEtran}
\bibliography{IEEEexample.bib}

\end{document}